\documentclass[11pt]{article}

\usepackage{acl2017}
\usepackage{times}
\usepackage{url}
\usepackage{booktabs}
\usepackage{latexsym}
\usepackage{adjustbox}
\usepackage[toc,page]{appendix}
\usepackage{graphicx}
\usepackage{color}
\usepackage{multirow}
\usepackage{enumitem}
\usepackage{tipa}
\usepackage{pifont}
\newcommand{\cmark}{\ding{51}}%
\newcommand{\xmark}{\ding{55}}%

\newcommand{\word}[1]{\texttt{#1}}
\newcommand{\feats}[1]{\textsf{#1}}
\newcommand{\affix}[1]{\texttt{#1}}
\newcommand{\gloss}[1]{`#1'}

\usepackage{tikz}

\usepackage[small]{caption}
\aclfinalcopy

\usepackage[disable]{todonotes}

\newcommand{\Jason}[2][]{\noindent}

\newcommand{\Ryan}[2][]{\noindent}
 \newcommand{\Mans}[2][]{\noindent}

\newcommand{\Geraldine}[2][]{\noindent}

\newcommand{\Chris}[2][]{\noindent}

\usepackage{cleveref}

\crefname{section}{\S}{\S\S}
\Crefname{section}{\S}{\S\S}
\crefname{table}{Table}{Tables}
\crefname{figure}{Figure}{Figures}
\crefname{algorithm}{Alg.}{Algs.}
\crefname{equation}{Eq.}{Eqs.}
\crefname{appendix}{Appendix}{Appendices}
\crefformat{section}{\S#2#1#3}

\newcommand{\ignore}[1]{}

\def\@fnsymbol#1{\ensuremath{\ifcase#1\or *\or \dagger\or \ddagger\or
   \mathsection\or \mathparagraph\or \|\or **\or \dagger\dagger
   \or \ddagger\ddagger \else\@ctrerr\fi}}

\title{CoNLL-SIGMORPHON 2017 Shared Task: \\ Universal Morphological Reinflection in 52 Languages}

\author{Ryan Cotterell$^{1}$ \and Christo Kirov$^1$ \and John Sylak-Glassman$^1$  \and \\ {\bf G{\'e}raldine Walther}$^2$ \and {\bf Ekaterina Vylomova$^{3}$} \and {\bf Patrick Xia$^{1}$} \and {\bf Manaal Faruqui}$^4$ \\ and {\bf Sandra K\"ubler}$^5$ \and {\bf David Yarowsky}$^1$ \and {\bf Jason Eisner}$^1$ \and {\bf Mans Hulden}$^{6}$ \\ Johns Hopkins University$^1$ \hspace{.1cm} University of Zurich$^2$ \hspace{.1cm} University of Melbourne$^3$ \\ Google$^4$ \hspace{.1cm} Indiana University$^5$ \hspace{.1cm} University of Colorado$^6$\\}
\date{}

\begin{document}\sloppy
\maketitle
\begin{abstract}
The CoNLL-SIGMORPHON 2017 shared task on supervised morphological generation required systems to be trained and tested in each of
52 typologically diverse languages. In sub-task 1,
submitted systems were asked to predict a specific inflected form of a given lemma.  In sub-task 2,
systems were given a lemma and some of its specific inflected forms, and asked to complete the inflectional paradigm by predicting all of the remaining inflected forms. Both sub-tasks included high, medium, and low-resource conditions. Sub-task 1 received 24 system submissions, while sub-task 2 received 3 system submissions. Following the success of neural sequence-to-sequence models in the SIGMORPHON 2016 shared task, all but one of the submissions included a neural component. The results show that high performance can be achieved with small training datasets, so long as models have appropriate inductive bias or make use of additional unlabeled data or synthetic data. However, different biasing and data augmentation resulted in non-identical sets of inflected forms being predicted correctly, suggesting that there is room for future improvement.

\end{abstract}

\section{Introduction}

Morphology interacts with both syntax and phonology.  As a result, explicitly modeling morphology has been shown to aid a number of tasks in human language technology (HLT), including machine translation (MT) \cite{dyer2008generalizing}, speech recognition \cite{creutz2007analysis}, parsing \cite{TACL631}, keyword spotting \cite{narasimhan2014morphological}, and word embedding \cite{CotterellSE16}.
Dedicated systems for modeling morphological patterns and complex word
forms have received less attention from the HLT community than tasks
that target other levels of linguistic structure.  Recently, however,
there has been a surge of work in this area
\cite{durrett2013supervised,ahlberg2014semi,nicolai2015inflection,faruqui2015morphological},
representing a renewed interest in morphology and the potential to use
advances in machine learning to attack a fundamental problem in
string-to-string transformations: the prediction of one
morphologically complex word form from another. This increased
interest in morphology as an independent set of problems within HLT
arrives at a particularly opportune time, as morphology is also
undergoing a methodological renewal within theoretical linguistics
where it is moving towards increased interdisciplinary work and
quantitative methodologies
\cite{moscosodelprado04,milin09,ackerman2009,sagot11sfcm,ackerman13,baayen13,blevins13infoth,pirrelli15,blevins16wp}. Pushing
the HLT research agenda forward in the domain of morphology promises
to lead to mutually highly beneficial dialogue between the two
fields.

Rich morphology is the norm among the languages of the world.  The linguistic typology database WALS shows that 80\% of the world's languages mark verb tense through morphology while 65\% mark grammatical case \cite{haspelmath2005world}.  The more limited inflectional system of English may help to explain the fact that morphology has received less attention in the computational literature than it is arguably due.

The CoNLL-SIGMORPHON 2017 shared task worked to promote the development of robust systems that can learn to perform cross-linguistically reliable morphological inflection and morphological paradigm cell filling using varying amounts of training data. We note that this is also the first CoNLL-hosted shared task to focus on morphology. The task itself featured training and development data from 52 languages representing a range of language families.  Many of the languages included were extremely low-resource, e.g., Quechua, Navajo, and Haida. The chosen languages also encompassed diverse morphological properties and inflection processes. Whenever possible, three data conditions were given for each language: low, medium, and high.  In the inflection sub-task, these corresponded to seeing 100 examples, 1,000 examples, and 10,000 examples respectively in the training data for almost all languages. The results show that encoder-decoder recurrent neural network models (RNNs) can perform very well even with small training sets, if they are augmented with various mechanisms to cope with the low-resource setting. The shared task training, development, and test data are released publicly.\footnote{\url{https://github.com/sigmorphon/conll2017}}

\begin{figure*}
\begin{center}
\includegraphics[width=0.74\textwidth]{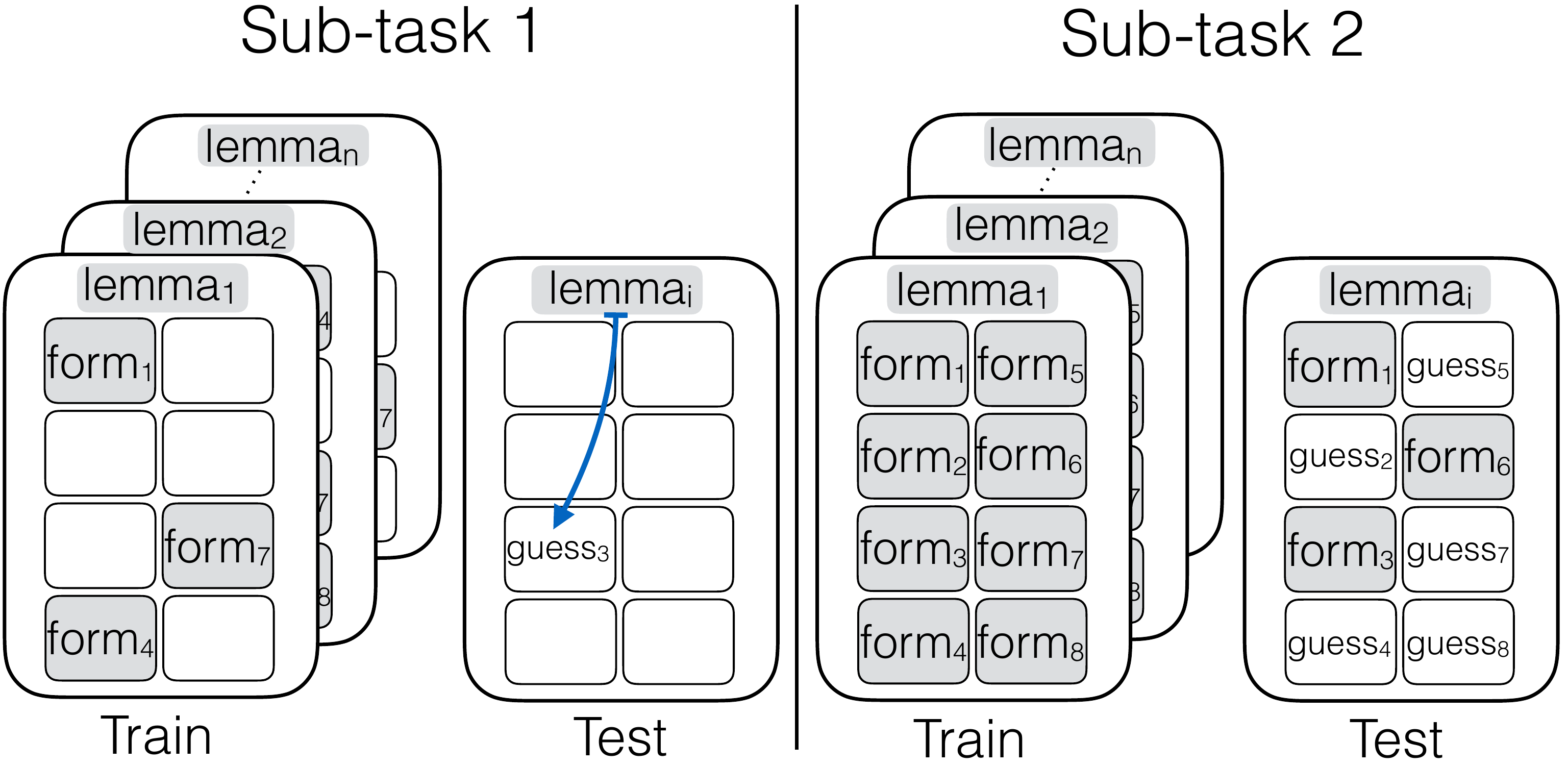}
\end{center}
\caption{Overview of sub-tasks.  Each large rectangle represents a \textbf{paradigm}, i.e., the full set of inflected forms for some lemma.  Each small rectangle within the paradigm is a \textbf{cell} that is associated with a known morphological feature bundle, and lists a string that either is observed (shaded background) or must be predicted (white background).  Sub-task 1 featured sparse training data and asked systems to inflect individual forms at test time.  Sub-task 2 provides dense paradigms as training data and asks for full paradigm completion of unseen items.}
\label{fig:taskoverview}
\end{figure*}

\begin{table}
\begin{adjustbox}{width=\columnwidth}
\begin{tabular}{llll}
\toprule
{\bf Lang} & {\bf Lemma}  & {\bf Inflection} & {\bf Inflected form} \\
\midrule
\multirow{ 2}{*}{en}    & \word{hug}              & \feats{V;PST}           & \word{hugged}           \\
                             & \word{spark}            & \feats{V;V.PTCP;PRS}    & \word{sparking}         \\
\midrule
\multirow{ 2}{*}{es}    & \word{liberar}          & \feats{V;IND;FUT;2;SG}  & \word{liberar\'as}        \\
                             & \word{descomponer}      & \feats{V;NEG;IMP;2;PL}  & \word{no descompong\'ais} \\
\midrule
\multirow{ 2}{*}{de}     & \word{aufbauen}         & \feats{V;IND;PRS;2;SG}  & \word{baust auf}        \\
                             & \word{\"Arztin}	          & \feats{N;DAT;PL}         & \word{\"Arztinnen}        \\
\bottomrule
\end{tabular}
\end{adjustbox}
\caption{Example training data from sub-task 1. Each training example maps a {\it lemma} and {\it inflection} to an {\it inflected form},
The inflection is a bundle of {\it morphosyntactic features}.  Note that inflected forms (and lemmata) can encompass multiple words. In the test data, the last column (the inflected form) must be predicted by the system.}
\label{tab:sub1data}
\end{table}
\section{Task and Evaluation Details}

This year's shared task contained two sub-tasks, which represented slightly different learning scenarios that might be faced by an HLT engineer or (roughly speaking) a human learner.
Beyond manually vetted\footnote{Thanks to: I\~naki Alegria, Gerlof Bouma, Zygmunt Frajzyngier, Chris Harvey, Ghazaleh Kazeminejad, Jordan Lachler, Luciana Marques, and Ruben Urizar.} data for training, development and test, monolingual corpus data (Wikipedia dumps) was also provided for both of the sub-tasks.  \Cref{fig:taskoverview} illustrates the two tasks and defines some terminology.

The CoNLL-SIGMORPHON 2017 shared task is the second shared task in a series that began with the SIGMORPHON 2016 shared task on morphological reinflection \cite{sigmorphon2016}.  In contrast to 2016, it happens that both of the 2017 sub-tasks actually involve only inflection, not reinflection.\footnote{\newcite{sigmorphon2016} defined the term: ``Systems developed for the 2016 Shared Task had to carry out {\em reinflection} of an already inflected form.  This involved {\em analysis} of an already inflected word form, together with {\em synthesis} of a different inflection of that form.''  In 2016, sub-task 1 involved only inflection while sub-tasks 2--3 required reinflection.}  Nonetheless, we kept ``reinflection'' in this year's title to make it easier to refer to the series of tasks.

\subsection{Sub-Task 1: Inflected Form from Lemma}\label{sec:subtask1}

\begin{table}
\begin{adjustbox}{width=\columnwidth}
\begin{tabular}{lll}
\toprule
\multicolumn{1}{c}{\bf Lemma}    & \multicolumn{1}{c}{\bf Inflections}
  &   \multicolumn{1}{c}{\bf Inflected forms}       \\
\midrule
\multicolumn{3}{c}{\textbf{Train}} \\
\midrule
\word{afrontar} &  \feats{V;IND;PST;1;PL;IPFV} &        \word{afront\'abamos} \\
\word{afrontar} &  \feats{V;SBJV;PST;3;PL;LGSPEC1} &        \word{afrontaran} \\
\word{afrontar} &  \feats{V;NEG;IMP;2;PL}  &        \word{no afront\'eis} \\
\word{afrontar} &  \feats{V;NEG;IMP;3;SG} &        \word{no afronte} \\
\word{afrontar} &  \feats{V;COND;2;SG}    &        \word{afrontar\'ias} \\
\word{afrontar} &  \feats{V;IND;FUT;3;SG} &        \word{afrontar\'a} \\
\word{afrontar} &  \feats{V;SBJV;FUT;3;PL} &        \word{afrontaren} \\
\multicolumn{3}{c}{\ldots} \\
\midrule
\multicolumn{3}{c}{\textbf{Test}} \\
\midrule
\word{revocar} & \feats{V;IND;PST;1;PL;IPFV}     &  \word{revoc\'abamos}          \\
\word{revocar} & \feats{V;SBJV;PST;3;PL;LGSPEC1} &  \multicolumn{1}{c}{---}      \\
\word{revocar} & \feats{V;NEG;IMP;2;PL}      &  \word{no revoqu\'eis}         \\
\word{revocar} & \feats{V;NEG;IMP;3;SG} &  \multicolumn{1}{c}{---}      \\
\word{revocar} & \feats{V;COND;2;SG} &  \word{revocar\'ias}           \\
\word{revocar} & \feats{V;IND;FUT;3;SG} &  \multicolumn{1}{c}{---}      \\
\word{revocar} & \feats{V;SBJV;FUT;3;PL} &  \multicolumn{1}{c}{---}      \\
\multicolumn{3}{c}{\ldots} \\
\bottomrule
\end{tabular}
\end{adjustbox}
\caption{Example training and test data from sub-task 2 in Spanish. At training time, the system is provided with complete paradigms, i.e., tables of all inflections for a given lemma, like the example at top.  At test time, the system is asked to complete partially filled paradigms, like the example at bottom; note that the inflectional features for the missing paradigm cells are provided in the input.}
\label{tab:sub2data}
\end{table}

The first sub-task in \cref{fig:taskoverview} required morphological generation with sparse training data, something that can be practically useful for MT and other downstream tasks in NLP.  Here, participants were given examples of inflected forms as shown in \cref{tab:sub1data}.  Each test example asked them to produce some other inflected form when given a lemma and a bundle of morphosyntactic features.

The training data was sparse in the sense that it included only a few inflected forms from each lemma.  That is, as in human L1 learning, the learner does not necessarily observe any complete paradigms in a language where the paradigms are large (e.g., dozens of inflected forms per lemma).\footnote{
Of course, human L1 learners do not get to observe explicit morphological feature bundles for the types that they observe.  Rather, they analyze inflected tokens in context
to discover both morphological features (including {\em inherent} features such as noun gender \cite{arnon12}) and paradigmatic structure (number of forms per lemma, number of expressed featural contrasts such as tense, number, person\ldots).
}

Key points:
\begin{enumerate}
 	\item Our sub-task 1 is similar to sub-task 1 of the SIGMORPHON 2016 shared task \cite{sigmorphon2016}, but with structured inflectional tags \cite{sylak-glassmankirov2015}, learning curve assessment, and many new typologically diverse languages, including low-resource languages.
	\item The task is inflection: Given an input lemma and desired output tag, participants had to generate the correct output inflected form (a string).
	\item The supervised training data consisted of individual forms (\cref{tab:sub1data}) that were sparsely sampled from a large number of paradigms.
        \item Forms that are empirically more frequent were more likely to appear in both training and test data (see \cref{sec:data} for details).
    \item Unannotated corpus data was also provided to participants.
	\item Systems were evaluated after training on $10^2$, $10^3$, and $10^4$ forms.
	\end{enumerate}

\subsection{Sub-Task 2: Paradigm Completion}\label{sec:paradigms}

The second sub-task in \cref{fig:taskoverview} focused on paradigm completion, also known as ``the paradigm cell filling problem'' \cite{ackerman2009}.

Here, participants were given a few complete inflectional paradigms as training data.  At test time, partially filled paradigms, i.e. paradigms with significant gaps in them, were to be completed by filling out the missing cells.  \Cref{tab:sub2data} gives examples.

Thus, sub-task 2 requires predicting many inflections of the same lemma.  Recall that sub-task 1 also required the system to predict several inflections of the same lemma (when they appear as separate examples in test data).  However, in sub-task 2, one of our test-time evaluation metrics (\cref{sec:evaluation}) is full-paradigm accuracy.  Also, the sub-task 2 training data provides full paradigms, in contrast to sub-task 1 where it included only a few inflected forms per lemma.  Finally, at test time, sub-task 2 presents each lemma along with some of its inflected forms, which is potentially helpful if the lemma had not appeared previously in training data.

Apart from the theoretical interest in this problem \cite{ackerman13}, this sub-task is grounded in the practical problem of extrapolation of basic resources for a language, where only a few complete paradigms may be available from a native speaker informant \cite{sylakglassmanremote} or a reference grammar.  L2 classroom instruction also asks human students to memorize example paradigms and generalize from them.

Key points:
  \begin{enumerate}
	\item The training data consisted of complete paradigms.
        \item Not all paradigms within a language have the same shape.  A noun lemma will have a different set of cells than a verb lemma does, and verbs of different classes (e.g., lexically perfective vs. imperfective) may also have different sets of cells.
	\item The task was paradigm completion: given a sparsely populated paradigm, participants should generate the inflected forms (strings) for all missing cells.
	\item The task simulates learning from compiled grammatical resources and inflection tables, or learning from a limited time with a native-language informant in a fieldwork scenario.
	\item Three training sets were given, building up in size from only a few complete paradigms to a large number (dozens).
\end{enumerate}

\subsection{Evaluation}\label{sec:evaluation}

Each team participating in a given sub-task was asked to submit $156$ versions of their system, where each version was trained using a different training set ($3$ training sizes $\times$ $52$ languages) and its corresponding development set.  We evaluated each submitted system on its corresponding test set, i.e., the test set for its language.

We computed three evaluation metrics: (i) Overall 1-best test-set accuracy, i.e., \textit{is the predicted paradigm cell correct?}  (ii) average Levenshtein distance, i.e., \textit{how badly does the predicted form disagree with the answer?} (iii) Full-paradigm accuracy, i.e., \textit{is the complete paradigm correct?}  This final metric only truly makes sense in sub-task 2, where full paradigms are given for evaluation. For each sub-task, the three data conditions (low, medium, and high) resulted in a learning curve.  For each system in each condition, we report the average metrics across all 52 languages.

\section{Data}\label{sec:data}

\subsection{Languages}

The data for the shared task was highly multilingual, comprising 52
unique languages.  Data for 47 of the languages came from the English edition of
Wiktionary, a large multi-lingual crowd-sourced dictionary containing morphological paradigms for many lemmata.\footnote{\url{https://en.wiktionary.org/}(08-2016 snapshot)} Data for
Khaling, Kurmanji Kurdish, and Sorani Kurdish was created as part of
the Alexina project
\cite{waltherjacques2013,walthersagot2010b,walthersagot2010}.\footnote{\url{https://gforge.inria.fr/projects/alexina/}}
Novel data for Haida, a severely endangered North American language
isolate, was prepared by Jordan Lachler (University of Alberta).  The
Basque language data was extracted from a manually designed
finite-state morphological analyzer \cite{alegria2009}.

The shared task language set is genealogically diverse, including languages
from 10 language stocks. Although the majority of the languages are
Indo-European, we also include two language isolates (Haida and
Basque) along with languages from Athabaskan (Navajo), Kartvelian
(Georgian), Quechua, Semitic (Arabic, Hebrew), Sino-Tibetan (Khaling),
Turkic (Turkish), and Uralic (Estonian, Finnish, Hungarian, and
Northern Sami). The shared task language set is also diverse in terms
of morphological structure, with languages which use primarily
prefixes (Navajo), suffixes (Quechua and Turkish), and a mix, with
Spanish exhibiting internal vowel variations along with suffixes and
Georgian using both infixes and suffixes. The language set also
exhibits features such as templatic morphology (Arabic, Hebrew), vowel
harmony (Turkish, Finnish, Hungarian), and consonant harmony (Navajo)
which require systems to learn non-local alternations. Finally, the
resource level of the languages in the shared task set varies greatly,
from major world languages (e.g.~Arabic, English, French, Spanish,
Russian) to languages with few speakers (e.g.~Haida, Khaling).

\subsection{Data Format}

For each language, the basic data consists of triples of the form
(lemma, feature bundle, inflected form), as in \cref{tab:sub1data}.
The first feature in the bundle always specifies the core part of
speech (e.g., verb).
  All features in the
bundle are coded according to the UniMorph Schema, a
cross-linguistically consistent universal morphological feature set
\cite{sylak-glassmankirov2015,sylakglassman-EtAl:2015:ACL-IJCNLP}.

\subsection{Extraction from Wiktionary}

For each of the 47 Wiktionary languages, Wiktionary provides a number
of tables, each of which specifies the full inflectional paradigm for
a particular lemma.  These tables were initially extracted via a
multi-dimensional table parsing strategy \cite{kirovsylak-glassman2016,sylak-glassmankirov2015}.

As noted in \cref{sec:paradigms}, different paradigms may have different shapes.
To prepare the shared task data, each language's
parsed tables from Wiktionary were grouped according to their tabular
structure and number of cells.  Each group represents a different type
of paradigm (e.g., verb).  We used only groups with a large number of
lemmata, relative to the number of lemmata available for the language
as a whole.
For each group, we associated a
feature bundle with each cell position in the table, by manually
replacing the prose labels describing grammatical features (e.g.~
``accusative case'') with UniMorph features (e.g.~\feats{acc}).
This allowed us to extract triples as described in the previous section.

By applying this process across the 47 languages, we constructed a large multilingual dataset that refines the parsed tables from previous work. This dataset was sampled to create appropriately-sized data for the shared task, as described in \cref{sec:sampling}.\footnote{Full, unsampled Wiktionary parses are made available at \url{unimorph.org} on a rolling basis.}
Full and sampled dataset sizes by language are given in \cref{tab:dq}.

Systematic syncretism is collapsed in Wiktionary.  For example, in English, feature bundles do not distinguish between different person/number forms of past tense verbs, because they are identical.\footnote{In this example, Wiktionary omits the single exception: the lemma \word{be} distinguishes between past tenses \word{was} and \word{were}.}  Thus, the past-tense form \word{went} appears only once in the table for \word{go}, not six times, and gives rise to only one triple, whose feature bundle specifies past tense but not person and number.

\subsection{Sampling the Train-Dev-Test Splits}\label{sec:sampling}

From each language's collection of paradigms, we sampled the training, development, and test sets as follows.  These datasets can be obtained from \url{http://www.sigmorphon.org/conll2017}.

Our first step was to construct probability distributions over the (lemma, feature bundle, inflected form) triples in our full dataset.  For each triple, we counted how many tokens the inflected form has in the February 2017 dump of Wikipedia for that language.  Note that this simple ``string match'' heuristic overestimates the count, since strings are ambiguous: not all of the counted tokens actually render that feature bundle.\footnote{For example, in English, any token of the string \word{walked} will be double-counted as both the past tense and the past participle of the lemma \word{walk}.
  This problem holds for all regular English verbs.  Similarly, when we are counting the present-tense tokens \word{lay} of the lemma \word{lay}, we will also include tokens of the string \word{lay} that are actually the past tense of \word{lie}, or are actually the adjective or noun senses of \word{lay}.  The alternative to double-counting each ambiguous token would have been to use EM to split the token's count of 1 unequally among its possible analyses, in proportion to their estimated prior probabilities \cite{TACL480}.}

From these counts, we estimated a unigram distribution over triples, using Laplace smoothing (add-1 smoothing).  We then sampled 12000 triples without replacement from this distribution.  The first 100 were taken as the low-resource training set for sub-task 1, the first 1000 as the medium-resource training set, and the first 10000 as the high-resource training set. Note that these training sets are nested, and that the highest-count triples tend to appear in the smaller training sets.

The final 2000 triples were randomly shuffled and then split in half to obtain development and test sets of 1000 forms each.  The final shuffling was performed to ensure that the development set is similar to the test set.  By contrast, the  development and test sets tend to contain lower-count triples than the training set.\footnote{This is a realistic setting, since supervised training is usually employed to generalize from frequent words that appear in annotated resources to less frequent words that do not.  Unsupervised learning methods also tend to generalize from more frequent words (which can be analyzed more easily by combining information from many contexts) to less frequent ones.}
In those languages where we have less than 12000 total forms, we omit the high-resource training set (all languages have at least 3000 forms).

To sample the data for sub-task 2, we perform a similar procedure.
For each paradigm in our full dataset, we counted the number of tokens in Wikipedia that matched any of the inflected forms in the paradigm.  From these counts, we estimated a unigram distribution over paradigms, using Laplace smoothing.
We sampled 300 paradigms without replacement from this distribution.
The low-resource training sets contain the first 10 paradigms, the medium-resource training set contains
the first 50, and high-resource training set contains the first 200. Again, these
training sets are nested. Note that since different
languages have paradigms of different sizes, the actual number of training exemplars may differ drastically.

With the same motivation as before, we shuffled the remaining 100 forms and
took the first 50 as development and the next 50 as test.  (In those languages with fewer than 300 forms, we again omitted the high-resource training setting.)
For each development or test paradigm, we chose about $\frac{1}{5}$ of the slots to provide to the system as input along with the lemma, asking the system to predict the remaining $\frac{4}{5}$.
We determined which cells to keep by independently flipping a biased coin with probability $0.2$ for each cell.

Because of the count overestimates mentioned above, our sub-task 1 dataset overrepresents triples where the inflected form (the answer) is ambiguous, and our sub-task 2 dataset overrepresents paradigms that contain ambiguous inflected forms.  The degree of ambiguity varied among languages: the average number of triples per inflected form string ranged from 1.00 in Sorani to 2.89 in Khaling, with an average of 1.43 across all languages.  Despite this distortion of true unigram counts, we believe that our datasets captured a sufficiently broad sample of the feature combinations for every language.

\begin{table*}
\centering
\begin{adjustbox}{width=.9\textwidth}
\begin{tabular}{l | l | r@{\ /\ }l | r r r | r | r || r r r}
\toprule
\textbf{Language} & \textbf{Family} & \textbf{Lemmata}&\textbf{Forms} & \textbf{High} & \textbf{Medium} & \textbf{Low} & \textbf{Dev} & \textbf{Test} & \textbf{Pr} & \textbf{Su} & \textbf{Ap} \\
\midrule
Albanian & Indo-European &589&33483&587 & 379  & 82  & 384 & 369               &56.24&95.14&1.09   \\
Arabic & Semitic&4134&140003 &3181  & 811  & 96  & 809  & 831               &54.64&90.89&31.61  \\
Armenian &Indo-European&7033&338461& 4657  & 907  & 99  & 875  & 902            &22.81&94.27&1.78   \\
Basque & Isolate &26&11889&26  & 26  & 22  & 26  & 22                        &97.63&92.07&12.87  \\
Bengali$\dagger$ & Indo-Aryan&136&4443&136  & 134  & 65  & 65  & 68                   &0.04&94.98&17.59   \\
Bulgarian & Slavic&2468&55730&2133  & 716  & 98  & 742  & 744            &15.65&92.09&4.28   \\
Catalan & Romance&1547&81576&1545  & 742  & 96  & 744  & 733                &0.41&98.04&6.89    \\
Czech & Slavic&5125&134527& 3862  & 836  & 98  & 852  & 850                &8.73&87.07&0.99    \\
Danish &Germanic&3193&25503& 3148  & 875  & 100  & 869  & 865              &0.17&81.52&1.28    \\
Dutch &Germanic&4993&55467& 4146  & 895  & 99  & 899  & 899                &3.06&80.61&4.30    \\
English &Germanic&22765&115523& 8377  & 989  & 100  & 985  & 983             &0.06&79.00&0.79    \\
Estonian &Uralic&886&38215& 886  & 587  & 94  & 553  & 577              &25.94&95.70&10.18  \\
Faroese & Germanic&3077&45474&2967  & 842  & 100  & 839  & 880               &0.66&80.52&12.93   \\
Finnish & Uralic&57642&2490377&8668  & 981  & 100  & 984  & 986            &31.47&94.47&10.57  \\
French &Romance&7535&367732& 5588  & 941  & 98  & 940  & 943               &2.79&97.78&3.95    \\
Georgian &Kartvelian&3782&74412& 3537  & 861  & 100  & 872  & 874            &3.28&94.70&0.42    \\
German &Germanic&15060&179339& 6767  & 959  & 100  & 964  & 964              &5.03&65.83&5.01    \\
Haida$\dagger$ &Isolate&41&7040& 41  & 41  & 40  & 34  & 38                       &0.26&98.96&0.49    \\
Hebrew &Semitic&510&13818& 510  & 470  & 95  & 431  & 453                &43.58&78.96&2.40   \\
Hindi &Indo-Aryan&258&54438& 258  & 252  & 85  & 254  & 255                &8.16&98.65&11.14   \\
Hungarian &Uralic&13989&490394& 7097  & 966  & 100  & 967  & 964              &0.52&97.00&0.52    \\
Icelandic &Germanic&4775&76915& 4108 & 899  & 100  & 906  & 899           &0.56&84.54&9.28    \\
Irish &Celtic&7464&107298& 5040  & 906  & 99  & 913  & 893                &55.09&61.60&4.47   \\
Italian &Romance&10009&509574& 6365  & 953  & 100  & 940  & 936             &18.81&92.38&20.92  \\
Khaling &Sino-Tibetan&591&156097& 584  & 426  & 92  & 411  & 422                &76.39&99.04&24.87  \\
Kurmanji Kurdish &Iranian&15083&216370& 7046  & 945  & 100  & 949  & 958   &9.62&91.43&0.90    \\
Latin &Romance&17214&509182& 6517  & 943  & 100  & 939  & 945              &4.12&90.04&47.74   \\
Latvian &Baltic&7548&136998& 5293  & 923  & 100  & 920  & 924            &3.69&91.50&2.91    \\
Lithuanian &Baltic&1458&34130& 1443  & 632  & 96  & 664  & 639           &3.64&90.58&35.32   \\
Lower Sorbian &Germanic &994&20121& 994  & 626  & 96  & 625  & 630         &0.24&93.33&0.48    \\
Macedonian &Slavic&10313&168057& 6079  & 958  & 100  & 939  & 946         &1.15&90.56&0.53    \\
Navajo &Athabaskan&674&12354& 674  & 496  & 91  & 491  & 491                  &79.03&35.08&21.49  \\
Northern Sami &Uralic&2103&62677& 1964  & 745  & 93  & 738  & 744        &4.62&90.39&18.12   \\
Norwegian Bokm{\aa}l &Germanic&5527&19238& 5041  & 925  & 100  & 928  & 930 &0.19&92.77&2.08    \\
Norwegian Nynorsk & Germanic&4689&15319&4413  & 915  & 98  & 914  & 919      &0.35&88.59&1.98    \\
Persian &Iranian&273&37128& 273  & 269  & 82  & 268  & 267               &27.1&95.28&15.70   \\
Polish &Slavic&10185&201024& 5926  & 929  & 99  & 934  & 942               &5.24&91.68&1.79    \\
Portuguese & Romance&4001&303996&3668  & 902  & 100  & 872  & 865         &0.01&93.26&3.19    \\
Quechua &Quechuan&1006&180004& 963  & 521  & 93  & 495  & 526              &1.25&98.92&0.05    \\
Romanian &Romance&4405&80266& 3351  & 858  & 99  & 854  & 828               &22.40&87.65&4.78   \\
Russian &Slavic&28068&473481& 8186  & 974  & 100  & 980  & 980            &5.20&79.88&11.33   \\
Scottish Gaelic$\dagger$ &Celtic&73&781& --- & 73  & 58  & 36  & 40                    &38.03&42.73&4.85   \\
Serbo-Croatian & Slavic&24419&840799&6746  & 964  & 100  & 971  & 954     &16.75&89.84&9.64   \\
Slovak &Slavic&1046&14796& 1046  & 631  & 93  & 622  & 622               &0.48&88.21&1.55    \\
Slovene &Slavic&2535&60110& 2007  & 769  & 100  & 746  & 762             &1.19&88.90&4.95    \\
Sorani Kurdish &Iranian&274&22990& 263  & 197  & 74  & 198  & 199       &67.89&94.76&15.21  \\
Spanish &Romance&5460&382955& 4621  & 906  & 99  & 902  & 922             &11.34&98.43&5.13   \\
Swedish &Germanic&10553&78411& 6511  & 962  & 99  & 956  & 962              &0.36&81.82&0.79    \\
Turkish &Turkic&3579&275460& 2934  & 834  & 99  & 852  & 840              &0.22&98.30&0.99    \\
Ukrainian &Slavic&1493&20904& 1490  & 722  & 98  & 744  & 729            &1.89&84.75&5.19    \\
Urdu &Indo-Aryan&182&12572& 182  & 111  & 55  & 101  & 106                 &8.01&95.93&8.10    \\
Welsh &Celtic&183&10641& 183  & 183  & 76  & 80  & 78                    &1.98&96.90&7.31    \\
\bottomrule
\end{tabular}
\end{adjustbox}
\caption{Total number of lemmata and forms available for sampling, and number of distinct lemmata present in each data condition in Task 1. For almost all languages, these were spread across 10000,1000, and 100 forms in the High, Medium, and Low conditions, respectively, and 1000 forms in each Dev and Test set. For $\dagger$-marked languages, there was not enough total data to support these numbers. Bengali had 4423 forms in the High condition, and Dev and Test sets of 100 forms each. Haida had 6840 forms in the High condition and Dev and Test sets of 100 forms. Scottish Gaelic had no High condition, a Medium condition of 681 forms, and Dev and Test sets of 50 forms each. The three last columns indicate how many inflected forms have undergone changes in a prefix (Pr), a change in a suffix (Su), or a stem-internal change (Ap) versus the given lemma form.}
\label{tab:dq}
\end{table*}

\begin{table}
\begin{adjustbox}{width=0.90\columnwidth}
\begin{tabular}{l  l l l} \toprule
Language Name & ADJ & N & V \\ \cline{1-1} \cline{2-4}
Albanian & -- & 10-20 & 123 \\
Arabic & 40-48 & 12-36 & 61-115 \\
Armenian & 17-34 & 17-34 & 154-155 \\
Basque & -- & -- & 112-810 \\
Bengali & -- & 9-12 & 51 \\
Bulgarian & 30 & 4-8 & -- \\
Catalan & -- & -- & 50-53 \\
Czech & 25-35 & 14 & 30 \\
Danish & -- & 6 & 8 \\
Dutch & 3-9 & -- & 16 \\
English & -- & -- & 7 \\
Estonian & -- & 30 & 79 \\
Faroese & 17 & 8-16 & 12 \\
Finnish & 28 & 13-28 & 141 \\
French & -- & -- & 49 \\
Georgian & 19 & 19 & -- \\
German & -- & 4-8 & 29 \\
Haida & -- & -- & 41-176 \\
Hebrew & -- & 30 & 23-28 \\
Hindi & -- & -- & 219 \\
Hungarian & -- & 17-34 & -- \\
Icelandic & -- & 8-16 & 28 \\
Irish & 13 & 7-13 & 65 \\
Italian & -- & -- & 47-51 \\
Khaling & -- & -- & 45-382 \\
Kurmanji Kurdish & 1-2 & 1-14 & 83 \\
Latin & 18-31 & 8-12 & 99 \\
Latvian & 20-24 & 7-14 & 49-50 \\
Lithuanian & 28-76 & 7-14 & 63 \\
Lower Sorbian & 33 & 18 & 21 \\
Macedonian & 16 & 5-11 & 20-29 \\
Navajo & -- & 8 & 6-50 \\
Northern Sami & 13 & 13 & 45-54 \\
Norwegian Bokm{\aa}l & 2-5 & 1-3 & 3-9 \\
Norwegian Nynorsk & 1-5 & 1-3 & 8 \\
Persian & -- & -- & 140 \\
Polish & 28 & 7-14 & 47 \\
Portuguese & -- & -- & 74-76 \\
Quechua & 256 & 256 & 41 \\
Romanian & 8-16 & 5-6 & 37 \\
Russian & 26-30 & 6-14 & 15-16 \\
Scottish Gaelic & 12 & -- & 8 \\
Serbo-Croatian & 1-43 & 2-14 & 63 \\
Slovak & 27 & 6-12 & -- \\
Slovene & 53 & 6-18 & 22 \\
Sorani Kurdish & 1-15 & 1-28 & 95-186 \\
Spanish & -- & -- & 70 \\
Swedish & 5-15 & 4-8 & 11 \\
Turkish & 72 & 12-108 & 120 \\
Ukrainian & 26 & 7-14 & 17-24 \\
Urdu & -- & 6 & 219 \\
Welsh & -- & -- & 20-65 \\ \bottomrule
\end{tabular}
\end{adjustbox}
\caption{Quantity of data available in sub-task 2. For each possible part of speech in each language, we present the range in the number of forms that comprise a paradigm as an indication of the difficulty of the task of forming a full paradigm. These ranges were computed using the data in the Train Medium condition.}
\end{table}

\section{Previous Work}
Most recent work in inflection generation has focused on sub-task 1,
i.e., generating inflected forms from the lemma.
Numerous, methodologically diverse approaches have been published. We highlight a representative sample of recent work. \newcite{durrett2013supervised} heuristically extracted
transformation rules and trained a semi-Markov model \cite{DBLP:conf/nips/SarawagiC04} to learn when to
apply them to the input. \newcite{nicolai2015inflection} trained a discriminative
string-to-string monotonic transduction tool---{\sc DirecTL+} \cite{jiampojamarn2008}---to
generate inflections. \newcite{ahlberg2014semi} reduced the problem to multi-class classification, where they used finite-state techniques to first generalize inflectional patterns and then trained a feature-rich classifier to choose the optimal such pattern to inflect unseen words \cite{ahlbergforsberg2015}. Finally,
\newcite{malouf2016}, \newcite{faruqui2015morphological} and
\newcite{kann-schutze:2016:P16-2} proposed a neural-based
sequence-to-sequence models \cite{DBLP:conf/nips/SutskeverVL14}, with \citeauthor{kann-schutze:2016:P16-2}
making use of an attention mechanism \cite{DBLP:journals/corr/BahdanauCB14}.
Overall, the neural approaches have generally been found to be the
most successful.

Some work has also focused on scenarios similar to sub-task 2.  For
example, \newcite{dreyer-eisner:2009:EMNLP} modeled the distribution
over the paradigms of a language as a Markov Random
Field (MRF), where each cell is represented as a
string-valued random variable.  The MRF's factors are specified as weighted
finite-state machines of the form given by \newcite{dreyer-smith-eisner:2008:EMNLP}. Building upon this,
\newcite{TACL480} proposed using a Bayesian network where both lemmata (repeated within a paradigm) and affixes (repeated across paradigms) were encoded as string-valued random variables.  That work required its finite-state transducers
to take a more restricted form \cite{cotterell-peng-eisner:2014:P14-2} for computational reasons. Finally, \newcite{kann-cotterell-schutze:2017:EACLlong}
proposed a multi-source sequence-to-sequence network, allowing
a neural transducer to exploit multiple source forms simultaneously.

\paragraph{SIGMORPHON 2016 Shared Task.}
Last year, the SIGMORPHON 2016 shared task (\url{http://sigmorphon.org/sharedtask}) focused on 10 languages (including 2 surprise languages). As for the present 2017 task, most of the 2016 data was derived from Wiktionary. The 2016 shared task had submissions from 9 competing teams with members from 11 universities.  As mentioned in \cref{sec:subtask1}, our sub-task 1 is an extension of sub-task 1 from 2016.  The other sub-tasks in 2016 focused on the more general reinflection problem, where systems had to learn to map from any inflected form to any other with varying degrees of annotations. See \newcite{sigmorphon2016} for details.

\section{The Baseline System}\label{subsec:baseline}

The shared task provided a baseline system to participants that
addressed both tasks and all languages.  The system was
designed for speed of application and also for adequate accuracy with
little training data, in particular in the low and medium data
conditions. The design of the baseline was inspired by the University of Colorado's
submission \cite{liu2016} to the SIGMORPHON 2016 shared task.

\subsection{Alignment}

For each (lemma, feature bundle, inflected form) triple in training data,
the system initially aligns the lemma with the inflected form by finding the minimum-cost edit path.  Costs are computed with a weighted scheme
such that substitutions have a slightly higher cost (1.1) than
insertions or deletions (1.0).  For example, the German training data
pair \word{schielen}-\word{geschielt} \gloss{to squint} (going from the
lemma to the past participle) is aligned as:

\begin{verbatim}
           --schielen
           geschielt-
\end{verbatim}

The system now assumes that each aligned pair can be broken up
into a prefix, stem and a suffix, based on
where the inputs or outputs have initial or trailing blanks after
alignment. We assume that initial or trailing blanks in either input
or output reflect boundaries between a prefix and a stem, or a stem
and a suffix.  This allows us to divide each training example into
three parts.  Using the example above, the pairs would be aligned as
follows, after padding the edges with \word{\$}-symbols:

\medskip

\begin{center}
\begin{tabular}{r|c|l}
\textbf{prefix} & \textbf{stem} & \textbf{suffix} \\
\affix{\$} & \affix{schiele}  & \affix{n\$} \\
\affix{\$ge} & \affix{schielt} & \affix{\$}   \\
\end{tabular}
\end{center}

\medskip

\subsection{Inflection Rules}

From this alignment, the system extracts a prefix-changing rule based on the prefix pairing, as well as a set of suffix-changing rules based on suffixes of the stem+suffix pairing.  The example alignment above yields the eight extracted suffix-modifying rules

\medskip

\begin{center}
\small
\begin{tabular}{ll}
\affix{n\$} $\rightarrow$ \affix{\$}
&
\affix{ielen\$} $\rightarrow$ \affix{ielt\$}
\\
\affix{en\$} $\rightarrow$ \affix{t\$}
&
\affix{hielen\$} $\rightarrow$ \affix{hielt\$}
\\
\affix{len\$} $\rightarrow$ \affix{lt\$}
&
\affix{chielen\$} $\rightarrow$ \affix{chielt\$}
\\
\affix{elen\$} $\rightarrow$ \affix{elt\$}
&
\affix{schielen\$} $\rightarrow$ \affix{schielt\$}
\\
\end{tabular}
\normalsize
\end{center}

\medskip

\noindent as well as the prefix-modifying rule \affix{\$} $\rightarrow$ \affix{\$ge}.

Since these rules were obtained from the triple (\word{schielen},
\feats{V;V.PTCP;PST}, \word{geschielt}), they are associated with
a token of the feature bundle \feats{V;V.PTCP;PST}.

\subsection{Generation}

\begin{table*}
  \begin{adjustbox}{width=2\columnwidth}
  \begin{tabular}{l l l } \toprule
    Team & Institute(s) & System Description Paper  \\ \midrule
CLUZH     & University of Zurich &  \newcite{makarov-ruzsics-clematide:2017:K17-20}  \\
CMU       & Carnegie Mellon University & \newcite{zhou-neubig:2017:K17-20} \\
CU        & University of Colorado Boulder & \newcite{silfverberg-EtAl:2017:K17-20} \\
EHU       & University of the Basque Country  & \newcite{ehu-shared-task}$^*$ \\
IIT (BHU) & Birla Institute of Technology and Science / & \newcite{sudhakar-singh:2017:K17-20} \\
& \hspace{0.5cm}Indian Institute of Technology (BHU) Varanasi  \\
ISI       & Indian Statistical Institute & \newcite{chakrabarty-garain:2017:K17-20} \\
LMU       & Ludwig-Maximilian University of Munich & \newcite{kann-schutze:2017:K17-20} \\
SU-RUG     & Stockholm University / University of Groningen & \newcite{ostling-bjerva:2017:K17-20} \\
UA        & University of Alberta & \newcite{nicolai-EtAl:2017:K17-20} \\
UE-LMU    & University of Edinburgh / & \newcite{bergmanis-EtAl:2017:K17-20} \\
 & \hspace{0.5cm} Ludwig-Maximilian University of Munich \\
UF        & University of Florida & \newcite{zhu-li-li:2017:K17-20} \\
UTNII     & National Institute of Informatics / & \newcite{senuma-aizawa:2017:K17-20} \\
& \hspace{0.5cm} University of Tokyo &  \\ \bottomrule
  \end{tabular}
  \end{adjustbox}
  \caption{The teams' abbreviations as well as their members' institutes and the accompanying system description paper are listed here. Note that in the main text the abbreviations are used with a integer index, indicating the specific submission. One team (marked $*$), did not submit a system description.}
  \label{tab:teams}
\end{table*}

At test time, to inflect a lemma with features, the baseline system applies rules associated with training tokens of the precise feature bundle.  There is no generalization across bundles that share features.

Specifically, the longest-matching suffix rule associated with the
feature bundle is consulted and applied to the input form.  Ties are
broken by frequency, in favor of the rule that has occurred most often
with this feature bundle.  After this, the prefix rule that occurred
most often with the bundle is likewise applied.  That is, the prefix-matching rule has no longest-match preference, while the suffix-matching rule does.

For example, to inflect \word{kaufen} \gloss{to buy} with the features \feats{V;V.PTCP;PST}, using the single example above as training data, we would find that the longest matching stored suffix-rule is \affix{en\$ $\rightarrow$ t\$}, which would transform \word{kaufen} into an intermediate form \word{kauft}, after which the most frequent prefix-rule, \affix{\$ $\rightarrow$ \$ge} would produce the final output \word{gekauft}.  If no rules have been associated with a particular feature bundle (as often happens in the low data condition), the inflected form is simply taken to be a copy of the lemma.

In sub-task 2, paradigm completion, the baseline system simply repeats the sub-task 1 method and generates all the missing forms independently from the lemma.  It does not take advantage of the other forms that are presented in the partially filled paradigm.

In addition to the above, the baseline system uses a heuristic to place a language
into one of two categories: largely prefixing or largely suffixing.
Some languages, such as Navajo, are largely prefixing and have more
complex changes in the left periphery of the input rather than at the
right. However, in the method described above, the operation of the prefix rules is more restricted than that
of the suffix rules: prefix rules tend to perform no
change at all, or insert or delete a prefix.  For largely
prefixing languages, the method performs better when operating with
reversed strings. Classifying a language into prefixing or suffixing
is done by simply counting how often there is a prefix change
vs. suffix change in going from the lemma form to the inflected form
in the training data. Whenever a language is found to be largely
prefixing, the system works with reversed strings throughout to allow
more expressive changes in the left edge of the input.

\section{System Descriptions}
The CoNLL-SIGMORPHON 2017 shared task received submissions from 11
teams with members from 15 universities and institutes (\cref{tab:teams}). Many of the
teams submitted more than one system, yielding a total of 25 unique
systems entered including the baseline system.

In contrast to the
2016 shared task, all but one of the submitted systems included
a neural component. Despite the relative uniformity of the submitted
architectures, we still observed large differences in the individual
performances. Rather than differences in architecture, a major
difference this year was the various methods for supplying the
neural network with auxiliary training data.  For ease of presentation,
we break down the systems into the features of their system (see
\cref{table:features}) and discuss the systems that had those features.
In all cases, further details of the methods can be found in the
system description papers, which are cited in \cref{tab:teams}.

\begin{table}
\begin{adjustbox}{width=\columnwidth}
\begin{tabular}{lccccc} \toprule
          &  Neural & Hard   & Rerank & Data$+$\\ \midrule
baseline  & \xmark  & \cmark & \xmark & \xmark \\
CLUZH     & \cmark  & \cmark & \xmark & \xmark  \\
CMU       & \cmark  & \xmark & \xmark & \cmark \\
CU        & \cmark  & \xmark & \xmark & \cmark \\
EHU       & \xmark  & \cmark & \xmark & \xmark \\
IIT (BHU) & \cmark  & \xmark & \xmark & \cmark \\
ISI       & \cmark  & \xmark & \cmark & \xmark \\
LMU       & \cmark  & \xmark & \xmark & \cmark \\
SU-RUG    & \cmark  & \xmark & \xmark & \xmark \\
UA        & \cmark  & \cmark & \cmark & \cmark \\
UE-LMU    & \cmark  & \xmark & \xmark & \cmark \\
UF        & \cmark  & \xmark & \xmark & \xmark \\
UTNII     & \cmark  & \xmark & \xmark & \xmark \\
\bottomrule
\end{tabular}
\end{adjustbox}
\caption{Features of the various submitted systems.}
\label{table:features}
\end{table}

\paragraph{Neural Parameterization.}
All systems except for the EHU team employed some form of a neural
network. Moreover, all teams except for SU-RUG, which employed a
convolutional neural network, made use of some form of gated recurrent
network---either a gated recurrent network (GRU) \cite{DBLP:journals/corr/ChungGCB14} or long
short-term memory (LSTM) \cite{DBLP:journals/neco/HochreiterS97}. In these neural models,
a common strategy was to feed in the morphological tag of the form to be predicted  along with the input into the network, where
each subtag was its own symbol.

\paragraph{Hard Alignment versus Soft Attention.}
Another axis, along which the systems differ is the use of hard
alignment, over soft attention. The neural attention mechanism was
introduced in \newcite{DBLP:journals/corr/BahdanauCB14} for neural machine translation (NMT). In short,
these mechanisms avoid the necessity of encoding the input word into a
fixed length vector, by allowing the decoder to attend to different
parts of the inputs. Just as in NMT, the attention mechanism has led
to large gains in morphological inflection. The CMU, CU, IIT (BHU),
LMU, UE-LMU, UF and UTNII systems all employed such mechanisms.

An alternative to soft attention is hard, monotonic alignment, i.e., a neural
parameterization of a traditional finite-state transduction
system. These systems enforce a monotonic alignment between source
and target forms. In the 2016 shared task \cite[see][Table
  6]{sigmorphon2016} such a system placed second
  \cite{aharoni-goldberg-belinkov:2016:SIGMORPHON},
and this year's winning system---CLUZH---was an extension of that one. (See, also,
\newcite{Aharoni} for a further explication of the technique and \newcite{rastogi-cotterell-eisner:2016:N16-1} for discussion of a related neural parameterization of a weighted finite-state machine.) Their
system allows for explicit biasing towards a copy action
that appears useful in the low-resource setting. Despite its neural
parameterization, the CLUZH system is most closely related to the systems of UA
and EHU, which train weighted finite-state transducers, albeit with a log-linear parameterization.

\paragraph{Reranking.}
Reranking the output of a weaker system was a tack taken by two
systems: ISI and UA.  The ISI system started with a heuristically
induced candidate set, using the edit tree approach described by
\newcite{CHRUPALA08.594}, and then chose the best edit tree. This
approach is effectively a neuralized version of the lemmatizer
proposed in \newcite{muller-EtAl:2015:EMNLP} and, indeed, was
originally intended for that task \cite{Chakrabarty}. The UA team,
following their 2016 submission, proposed a linear reranking on top of
the $k$-best output
of their transduction system.

\paragraph{Data Augmentation.}
Many teams made use of auxiliary training data---unlabeled or synthetic forms. Some teams leveraged the provided Wikipedia corpora (see \cref{sec:data}). The UE-LMU team used these unlabeled corpora to bias their methods towards copying by transducing an unlabeled word to itself.  The same team also explored a similar setup that instead learned to transduce {\em random} strings to themselves, and found that using random strings worked almost as well as words that appeared in unlabeled corpora.
CMU used a variational autoencoder and treated the tags of unannotated words in the Wikipedia corpus as latent variables (see \newcite{multi-space} for more details).
Other teams attempted to get silver-standard labels for the unlabeled
corpora.  For example, the UA team trained a tagger on the given
training examples, and then tagged the corpus with the goal to obtain
additional instances, while the UE-LMU team used a series of unsupervised
heuristics.
The CU team---which did not make use of external resources---hallucinated more training data by identifying suffix and prefix changes in the given training pairs and then using that information to create new artificial training pairs.
The LMU submission also experimented with hand-written rules
to artificially
generate more data. It seems likely that the primary difference in the performance
of the various neural systems lay in these strategies for the creation of new data to train the parameters, rather than in the neural architectures themselves.

\begin{table}
\centering
\begin{adjustbox}{width=.95\columnwidth}
\begin{tabular}{llll}
\toprule
& \multicolumn{1}{c}{High} & \multicolumn{1}{c}{Medium} & \multicolumn{1}{c}{Low} \\
\midrule
UE-LMU-1 & \textbf{95.32/0.10} & 81.02/0.41 & \hspace{14pt}---/---\\
CLUZH-7 & 95.12/0.10 & \textbf{82.80/0.34} & \textbf{50.61/1.29}\\
CLUZH-6 & 95.12/0.10 & \textbf{82.80/0.34} & \textbf{50.61/1.29}\\
CLUZH-2 & 94.95/0.10 & 81.80/0.37 & 46.82/1.38\\
LMU-2 & 94.70/0.11 & 82.64/0.35 & 46.59/1.56\\
LMU-1 & 94.70/0.11 & 82.64/0.35 & 45.29/1.62\\
CLUZH-5 & 94.69/0.11 & 81.00/0.39 & 48.24/1.48\\
CLUZH-1 & 94.47/0.12 & 80.88/0.39 & 45.99/1.43\\
SU-RUG-1 & 93.56/0.14 & \hspace{14pt}---/--- & \hspace{14pt}---/---\\
CU-1 & 92.97/0.17 & 77.60/0.50 & 45.74/1.62\\
UTNII-1 & 91.46/0.17 & 65.06/0.73 & 1.28/5.71\\
CLUZH-4 & 89.53/0.23 & 80.33/0.41 & 48.53/1.52\\
IIT(BHU)-1 & 89.38/0.22 & 50.73/1.69 & 13.88/4.54\\
CLUZH-3 & 89.10/0.24 & 79.57/0.44 & 47.95/1.55\\
UF-1 & 87.33/0.27 & 68.82/0.78 & 27.46/2.70\\
CMU-1$\dagger$ & 86.56/0.28 & 68.00/0.86$\ddagger$ & \hspace{14pt}---/---\\
ISI-1 & 74.01/0.78 & 54.47/1.39 & 26.00/2.43\\
EHU-1 & 64.38/0.72$\ddagger$ & 38.50/1.70$\ddagger$ & 3.50/3.23$\ddagger$\\
UE-LMU-2$\dagger$ & \hspace{14pt}---/--- & 82.37/0.39 & \hspace{14pt}---/---\\
IIT(BHU)-2 & \hspace{14pt}---/--- & 55.46/1.78 & 14.27/4.33\\
UA-3$\dagger$ & \hspace{14pt}---/--- & \hspace{14pt}---/--- & 57.70/1.34$\ddagger$\\
UA-4$\dagger$ & \hspace{14pt}---/--- & \hspace{14pt}---/--- & 57.52/1.36$\ddagger$\\
UA-1 & \hspace{14pt}---/--- & \hspace{14pt}---/--- & 54.22/1.66$\ddagger$\\
UA-2 & \hspace{14pt}---/--- & \hspace{14pt}---/--- & 42.85/2.23$\ddagger$\\
\midrule
baseline & 77.81/0.50 & 64.70/0.90 & 37.90/2.15\\
\midrule
oracle-fc & 99.99/* & 97.76/* & 70.84/*\\
oracle-e & 98.25/* & 92.10/* & 64.56/*\\
\bottomrule
\end{tabular}
\end{adjustbox}
\setlength\belowcaptionskip{-5pt}
\caption{Sub-task 1 results: Per-form accuracy (in \%age points) and average Levenshtein distance from the correct form (in characters), averaged across the 52 languages with all languages weighted equally. The columns represent the different training size conditions.  Systems marked with $\dagger$ used external resources. Accuracies marked with $\ddagger$ indicate that the submission did not include all 52 languages and should not be compared to the other accuracies.}\label{tab:t1perfsum}
\end{table}

\begin{table}
\centering
\begin{tabular}{llll}
\toprule
& \multicolumn{1}{c}{High} & \multicolumn{1}{c}{Medium} & \multicolumn{1}{c}{Low} \\
\midrule
LMU-2 & \textbf{88.52/0.22} & \textbf{82.02/0.38} & \textbf{67.76/0.75}\\
LMU-1 & 87.40/0.24 & 77.02/0.47 & 54.74/1.22\\
CU-1 & 67.77/0.75 & 60.94/1.03 & 47.89/1.67\\
\midrule
baseline & 76.87/0.51 & 65.84/0.83 & 50.14/1.28\\
\midrule
oracle-e & 94.11/* & 88.70/* & 75.84/*\\
\bottomrule
\end{tabular}
\caption{Sub-task 2 results: Per-form accuracy (in \%age points) and average Levenshtein distance from the correct form (in characters).}\label{tab:t2perfsum}
\end{table}

\begin{table*}
\begin{adjustbox}{width=\textwidth}
\begin{tabular}{lllllll}
\toprule
& \multicolumn{3}{c}{Sub-task 1} &\multicolumn{3}{c}{Sub-task 2}\\
& High & Medium & Low & High & Medium & Low\\
\midrule
Albanian & 99.00(UE-LMU) & 89.40(CU-1) & 31.00(CU-1) & 98.35(LMU-2) & 88.81(LMU-1) & 66.63(LMU-2)\\
Arabic & 94.50(CLUZH-7) & 79.70(LMU-2) & 37.00(CLUZH-7) & 95.48(LMU-2) & 90.21(LMU-2) & 80.43(LMU-2)\\
Armenian & 97.50(UE-LMU) & 91.50(LMU-2) & 58.70(CLUZH-7) & 98.78(LMU-2) & 97.77(LMU-2) & 93.92(LMU-2)\\
Basque & 100.00(UTNII-1) & 89.00(UE-LMU) & 20.00(LMU-2) & --- & 94.14(LMU-2) & 93.02(LMU-2)\\
Bengali & 100.00(UE-LMU) & 99.00(CLUZH-1) & 68.00(CLUZH-3) & 92.61(LMU-1) & 91.72(LMU-2) & 90.19(LMU-2)\\
Bulgarian & 98.10(UE-LMU) & 82.50(LMU-2) & 57.10(CU-1) & 85.93(LMU-2) & 55.95(LMU-2) & 49.58(LMU-2)\\
Catalan & 98.40(CLUZH-1) & 92.60(CLUZH-7) & 66.40(CU-1) & 99.35(LMU-2) & 97.06(LMU-2) & 94.16(baseline)\\
Czech & 94.10(UE-LMU) & 86.30(CU-1) & 44.00(CLUZH-7) & 86.00(LMU-1) & 58.61(LMU-2) & 34.96(LMU-2)\\
Danish & 94.50(UE-LMU) & 83.60(LMU-2) & 75.50(CLUZH-7) & 75.74(LMU-2) & 71.15(baseline) & 53.11(CU-1)\\
Dutch & 96.90(UE-LMU) & 86.50(LMU-2) & 53.60(baseline) & 89.30(LMU-2) & 86.53(LMU-2) & 56.64(LMU-2)\\
English & 97.20(UE-LMU) & 94.70(LMU-2) & 90.60(UA-1) & 91.60(baseline) & 84.00(baseline) & 84.40(CU-1)\\
Estonian & 98.90(UE-LMU) & 82.40(UE-LMU) & 32.90(CLUZH-7) & 97.90(LMU-2) & 92.43(LMU-2) & 77.42(LMU-2)\\
Faroese & 87.80(CLUZH-7) & 68.10(CLUZH-7) & 42.40(CLUZH-7) & 71.90(LMU-2) & 68.31(LMU-2) & 57.55(LMU-2)\\
Finnish & 95.10(UE-LMU) & 78.40(UE-LMU) & 19.70(CLUZH-7) & 93.67(LMU-2) & 89.48(LMU-2) & 76.30(LMU-2)\\
French & 89.50(UE-LMU) & 80.30(CLUZH-7) & 66.00(CLUZH-7) & 98.83(LMU-2) & 95.38(LMU-2) & 87.45(LMU-2)\\
Georgian & 99.40(LMU-2) & 93.40(CLUZH-7) & 85.60(LMU-2) & 96.20(LMU-2) & 89.67(LMU-2) & 86.82(LMU-2)\\
German & 93.00(UE-LMU) & 80.00(CLUZH-4) & 68.10(CLUZH-4) & 85.88(LMU-2) & 77.56(LMU-2) & 74.66(LMU-2)\\
Haida & 99.00(UTNII-1) & 95.00(LMU-2) & 46.00(LMU-2) & --- & 96.40(LMU-2) & 95.24(LMU-2)\\
Hebrew & 99.50(LMU-2) & 83.80(LMU-2) & 35.40(CU-1) & 93.42(LMU-2) & 85.59(LMU-2) & 68.06(LMU-2)\\
Hindi & 100.00(UTNII-1) & 97.40(CLUZH-3) & 75.50(LMU-2) & 99.95(LMU-2) & 95.01(LMU-2) & 93.84(LMU-2)\\
Hungarian & 86.80(CLUZH-7) & 75.10(CLUZH-4) & 38.10(CLUZH-7) & 89.04(LMU-2) & 79.97(LMU-2) & 54.50(LMU-2)\\
Icelandic & 92.10(CLUZH-7) & 74.70(CLUZH-7) & 40.80(CU-1) & 74.30(LMU-1) & 67.21(LMU-2) & 56.57(LMU-2)\\
Irish & 92.10(CLUZH-7) & 72.60(CLUZH-7) & 37.80(CLUZH-7) & 69.53(LMU-2) & 52.92(LMU-2) & 43.43(LMU-2)\\
Italian & 97.90(UE-LMU) & 93.30(UE-LMU) & 56.40(CU-1) & 97.05(LMU-2) & 90.67(LMU-2) & 72.00(LMU-2)\\
Khaling & 99.50(UE-LMU) & 87.10(LMU-2) & 18.00(LMU-2) & 99.73(LMU-2) & 98.62(LMU-2) & 97.15(LMU-2)\\
Kurmanji & 94.80(UE-LMU) & 92.80(CLUZH-7) & 86.60(CLUZH-2) & 94.26(LMU-2) & 88.87(LMU-2) & 80.17(LMU-2)\\
Latin & 81.30(UE-LMU) & 51.80(CLUZH-7) & 19.30(CU-1) & 87.70(LMU-2) & 84.63(LMU-2) & 51.98(LMU-2)\\
Latvian & 97.30(UE-LMU) & 88.60(CLUZH-7) & 68.10(CLUZH-4) & 96.69(LMU-2) & 89.19(LMU-2) & 75.79(LMU-2)\\
Lithuanian & 95.80(UE-LMU) & 62.60(UE-LMU) & 23.30(baseline) & 85.82(LMU-2) & 82.87(LMU-2) & 49.51(LMU-2)\\
Lower Sorbian & 97.50(UE-LMU) & 84.10(UE-LMU) & 52.30(CU-1) & 87.39(LMU-2) & 84.02(LMU-2) & 56.43(LMU-2)\\
Macedonian & 97.30(UE-LMU) & 91.80(CLUZH-1) & 65.50(CLUZH-7) & 97.14(LMU-2) & 88.98(LMU-2) & 60.23(LMU-2)\\
Navajo & 92.30(UE-LMU) & 50.80(CLUZH-7) & 20.40(CLUZH-7) & 58.22(LMU-2) & 47.12(LMU-2) & 35.48(LMU-2)\\
Northern Sami & 98.60(UE-LMU) & 74.00(UE-LMU) & 18.70(CU-1) & 91.56(LMU-2) & 83.51(LMU-2) & 39.86(LMU-2)\\
Norwegian Bokm{\aa}l & 92.60(CLUZH-2) & 84.40(UE-LMU) & 78.00(CLUZH-7) & 70.44(CU-1) & 57.23(CU-1) & 49.06(CU-1)\\
Norwegian Nynorsk & 92.80(CLUZH-1) & 65.60(LMU-2) & 54.60(CLUZH-7) & 64.42(baseline) & 60.74(baseline) & 42.33(baseline)\\
Persian & 99.90(LMU-2) & 91.90(UE-LMU) & 51.00(CLUZH-7) & 100.00(LMU-2) & 99.56(LMU-2) & 99.20(LMU-2)\\
Polish & 92.80(UE-LMU) & 79.90(CLUZH-7) & 47.90(CLUZH-7) & 90.27(baseline) & 82.71(LMU-2) & 64.53(LMU-2)\\
Portuguese & 99.30(LMU-2) & 95.00(LMU-2) & 73.30(CLUZH-7) & 98.84(LMU-1) & 98.58(LMU-2) & 96.94(LMU-2)\\
Quechua & 100.00(CLUZH-4) & 98.30(CLUZH-7) & 61.10(CLUZH-7) & 99.84(LMU-2) & 99.60(LMU-2) & 99.98(LMU-2)\\
Romanian & 89.10(UE-LMU) & 77.40(CU-1) & 46.30(CLUZH-7) & 78.99(baseline) & 76.63(LMU-2) & 25.00(LMU-2)\\
Russian & 92.80(CLUZH-2) & 84.10(CLUZH-2) & 52.30(CLUZH-7) & 87.42(CU-1) & 85.74(LMU-2) & 46.17(LMU-2)\\
Scottish Gaelic & --- & 90.00(UE-LMU) & 64.00(CLUZH-3) & --- & 51.82(LMU-1) & 50.61(LMU-2)\\
Serbo-Croatian & 93.80(CLUZH-2) & 83.30(CU-1) & 39.20(CU-1) & 88.29(LMU-2) & 59.18(LMU-2) & 40.46(LMU-2)\\
Slovak & 95.30(CLUZH-2) & 80.50(CLUZH-7) & 53.60(CLUZH-7) & 71.84(LMU-2) & 66.67(LMU-2) & 53.65(LMU-2)\\
Slovene & 97.10(CLUZH-5) & 88.80(LMU-2) & 63.00(CLUZH-7) & 93.71(LMU-1) & 85.10(LMU-2) & 79.28(LMU-2)\\
Sorani & 89.40(CLUZH-7) & 82.90(LMU-2) & 27.10(CU-1) & 86.39(LMU-2) & 86.05(LMU-2) & 57.65(LMU-2)\\
Spanish & 97.50(CLUZH-7) & 91.70(UE-LMU) & 66.40(CLUZH-7) & 98.53(LMU-2) & 97.89(LMU-2) & 91.05(LMU-2)\\
Swedish & 93.10(UE-LMU) & 79.70(UE-LMU) & 64.20(CLUZH-3) & 84.71(LMU-2) & 70.88(LMU-2) & 51.18(LMU-2)\\
Turkish & 98.40(UE-LMU) & 89.70(UE-LMU) & 42.00(CLUZH-7) & 99.41(LMU-2) & 98.65(LMU-2) & 87.65(LMU-2)\\
Ukrainian & 95.00(UE-LMU) & 82.50(CLUZH-7) & 50.40(CU-1) & 74.76(LMU-1) & 67.14(baseline) & 49.21(LMU-2)\\
Urdu & 99.70(UE-LMU) & 98.00(CLUZH-4) & 74.10(CLUZH-7) & 98.44(LMU-1) & 94.29(LMU-2) & 88.53(LMU-2)\\
Welsh & 99.00(CLUZH-1) & 93.00(LMU-2) & 56.00(CLUZH-7) & 97.96(LMU-2) & 97.80(LMU-2) & 89.89(LMU-2)\\
\bottomrule
\end{tabular}
\end{adjustbox}
\caption{Best per-form accuracy (and corresponding system) by language.}\label{tab:bestbylanguage}
\end{table*}

\section{Performance of the Systems}

Relative system performance is described in \cref{tab:t1perfsum,tab:t2perfsum}, which show the average per-language accuracy of
each system by resource condition, for each of the sub-tasks. The table reflects the fact that some
teams submitted more than one system (e.g.~LMU-1 \& LMU-2 in the
table). Learning curves for each language across conditions are shown in \cref{tab:bestbylanguage}, which indicates the best per-form accuracy achieved by a submitted system. Full results can be found in \cref{appends}, including full-paradigm accuracy.

Three teams exploited external resources in some form: UA, CMU, and UE-LMU. In general, any relative performance gained was minimal. The CMU system was outranked by several systems that avoided external resource use in the High and Medium conditions in which it competed. UE-LMU only submitted a system that used additional resources in the Medium condition, and saw gains of $\sim$\%1 compared to their basic system, while it was still outranked overall by CLUZH. In the Low condition, UA saw gains of $\sim$\%3 using external data. However, all UA submissions were limited to a small handful of languages.

All but one of the systems submitted were neural. As expected given the
results from SIGMORPHON 2016, these systems perform very well when in
the High training condition where data is relatively plentiful. In the
Low and Medium conditions, however, standard encoder-decoder
architectures perform worse than the baseline using only the
training data provided. Teams that beat the baseline succeeded by
biasing networks towards the correct solutions through pre-training on
synthetic data designed to capture the overall inflectional patterns
in a language. As seen in \cref{tab:bestbylanguage}, these techniques worked better for some languages than for others. Languages with smaller, more regular paradigms were handled well (e.g., English sub-task 1 low-resource accuracy was at 90\%). Languages with more complex systems, like Latin, proved more challenging (the best system  achieved only 19\% accuracy in the low condition). For these languages, it is possible that the relevant variation required to learn a best per-form inflectional pattern was simply not present in the limited training data, and that a language-specific learning bias was required.

Even though the top-ranked systems do well on their own, different
systems may contain some amount of complementary information, so
that an ensemble over multiple approaches has a chance to improve
accuracy. We present an upper bound on the possible performance of
such an ensemble. \cref{tab:t1perfsum} and \cref{tab:t2perfsum}
include an ``Ensemble Oracle'' system (oracle-e) that gives the correct
answer if \emph{any} of the submitted systems is correct. The oracle
performs significantly better than any one system in both the Medium
($\sim$10\%) and Low ($\sim$15\%) conditions. This suggests that the
different strategies used by teams to ``bias'' their systems in an effort
to make up for sparse data lead to substantially different
generalization patterns.

For sub-task 1, we also present a second ``Feature Combination''
Oracle (oracle-fc) that gives the correct answer for a given test
triple iff its feature bundle appeared in training (with any lemma).
Thus, oracle-fc provides an upper bound on the performance of systems
that treat a feature bundle such as \feats{V;SBJV;FUT;3;PL} as atomic.
In the low-data condition, this upper bound was only 71\%, meaning
that 29\% of the test bundles had never been seen in training data.
Nonetheless, systems should be able to make some accurate predictions
on this 29\% by decomposing each test bundle into individual
morphological features such as \feats{FUT} (future) and \feats{PL}
(plural), and generalizing from training examples that involve
those features.
  For example, a particular feature or
sub-bundle might be realized as a particular affix.
 Several of the systems treated each individual feature as a separate input to the recurrent network, in order to enable this type of generalization. In the medium data condition for some languages, these systems sometimes far surpassed oracle-fc.  The most notable example of this is Basque, where oracle-fc produced a 47\% accuracy while six of the submitted systems produced an accuracy of 85\% or above. Basque is an extreme example with very large paradigms for the verbs that inflect in the language (only a few dozen common ones do).  This result demonstrates the ability of the neural systems to generalize and correctly inflect according to unseen feature combinations.

\section{Future Directions}
As regards morphological inflection, there is a plethora of future
directions to consider. First, one might consider
morphological transductions over pronunciations, rather than
spellings.  This is more challenging in the many languages
(including English) where the orthography does not reflect the
phonological changes that accompany morphological processes
such as affixation.  Orthography usually also does not reflect
predictable allophonic distinctions in pronunciation \cite{sampson1985}, which one might
attempt to predict, such as the difference in aspiration
of /t/ in English [\textipa{t\super
    h}\textipa{A}p] (\word{top}) vs. [st\textipa{A}p] (\word{stop}).

A second future direction involves the effective
incorporation of external unannotated monolingual corpora into the
state-of-the-art inflection or reinflection systems. The best systems
in our competition {\em did not} make use of external data and those
that did make heavy use of such data, e.g., the CMU team, did not see
much gain.The best way to use external corpora remains an open
question; we surmise that they can be useful, especially in the
lower-resource cases. A related line of inquiry is the incorporation
of cross-lingual information, which
\newcite{kann-cotterell-schutze:2017:P17} did find to be helpful.

A third direction revolves around the efficient elicitation of
morphological information (i.e., active learning). In the low-resource section, we asked our
participants to find the best approach to generate new forms given
existing morphological annotation. However, it remains an open
question, which of the cells in a paradigm are best to collect
annotation for in the first place. Likely, it is better to collect {\em diagnostic} forms that are closer
to principal parts
of the paradigm \cite{finkel2007principal,ackerman2009,montermini13,cotterell-sylakglassman-kirov:2017:EACLshort}as these
will contain enough information such that the remaining
transformations are largely deterministic.
Experimental studies however suggest that speakers also strongly rely on pattern frequencies for inferring unknown forms \cite{seyfarth14}. Another interesting direction would therefore also include the organization of data according to plausible real frequency distributions (especially in spoken data) and exploring possibly varying learning strategies associated with lexical items of various frequencies. 

Finally, there is a wide variety of other tasks involving
morphology. While some of these have had a shared task, e.g., the
parsing of morphologically-rich languages
\cite{tsarfaty2010statistical} and unsupervised morphological
segmentation \cite{kurimo2010morpho}, many have not, e.g., supervised
morphological segmentation and morphological tagging. A key purpose of
shared tasks in the NLP community is the preparation and release of
standardized data sets for fair comparison among methods. Future
shared tasks in other areas of computational morphology would seem in
order, giving the overall effectiveness of shared tasks in unifying
research objectives in subfields of NLP, and as a starting point for possible cross-over with cognitively-grounded theoretical and quantitative linguistics.

\section{Conclusion}

The CoNLL-SIGMORPHON shared task provided an evaluation on 52 languages, with large and small datasets, of systems for
inflection and paradigm completion---two core tasks in computational morphological learning.  On sub-task 1 (inflection), 24 systems were submitted, while on sub-task 2 (paradigm completion), 3 systems were submitted. All but one of the systems used rather similar neural network models, popularized by the SIGMORPHON shared task in 2016.

The results reinforce the conclusions of the 2016 shared task that
encoder-decoder architectures perform strongly when training data
is plentiful, with exact-match accuracy on held-out forms surpassing 90\% on many
languages; we note there was a shortage of non-neural systems this year to compare with.
 In addition, and contrary to common expectation, many participants showed that neural systems can do reasonably well even with small training datasets.  A baseline sequence-to-sequence model achieves close to zero accuracy: e.g., \newcite{silfverberg-EtAl:2017:K17-20} reported that all the team's neural models on the low data condition delivered accuracies in the 0-1\% range without data augmentation, and other teams reported similar findings.
However, with judicious application of biasing and data augmentation techniques, the best neural systems achieved over 50\% exact-match prediction of inflected form strings on 100 examples, and 80\% on 1,000 examples, as compared to 38\% for a baseline system that learns simple inflectional rules.  It is hard to say whether these are ``good'' results in an absolute sense.  An interesting experiment would be to pit the small-data systems against human linguists who do not know the languages, to see whether the systems are able to identify the predictive patterns that humans discover (or miss).

An oracle ensembling of all systems shows that there is still much room for improvement, in particular in low-resource settings. We have released the training, development, and test sets, and expect these datasets to provide a useful benchmark for future research into learning of inflectional morphology and string-to-string transduction.

\section*{Acknowledgements}

The first author would like to acknowledge the support of an NDSEG fellowship.  Google provided support for the shared task in the form of an award. Several authors (CK, DY, JSG, MH) were supported in part by the Defense Advanced Research Projects Agency (DARPA) in the program Low Resource Languages for Emergent Incidents (LORELEI) under contract No. HR0011-15-C-0113. Any opinions, findings and conclusions or recommendations expressed in this material are those of the authors and do not necessarily reflect the views of the Defense Advanced Research Projects Agency (DARPA).

\bibliographystyle{acl_natbib}
\bibliography{conll_proposal}

\end{document}